\documentclass[runningheads]{llncs}

\usepackage{eccvabbrv}

\usepackage{graphicx}
\usepackage{booktabs}

\usepackage[accsupp]{axessibility}  % Improves PDF readability for those with disabilities.

\usepackage{hyperref}

\usepackage{orcidlink}

\begin{document}

\title{MMArt: A Multi-Perspective Multimodal Dataset for Visual Art Understanding} 

\titlerunning{Abbreviated paper title}

\author{
Shuai Wang\inst{1}\orcidlink{0000-0002-1595-3619} \and
Wangyuan Ding\inst{1}\orcidlink{0000-0002-9575-108X} \and
Yixian Shen\inst{1}\orcidlink{0000-0001-8447-872X} \and
Jia-Hong Huang\inst{1,2}\orcidlink{0000-0001-7943-2591} \and
Stevan Rudinac\inst{1}\orcidlink{0000-0003-1904-8736} \and
Monika Kackovic\inst{1}\orcidlink{0000-0002-7423-3902} \and
Nachoem Wijnberg\inst{1,3}\orcidlink{0000-0001-8070-8719} \and
Marcel Worring\inst{1}\orcidlink{0000-0003-4097-4136}
}

\authorrunning{S.~Wang et al.}

\institute{
University of Amsterdam, Amsterdam, The Netherlands\\
\email{\{s.wang3,w.ding,y.shen,j.huang,s.rudinac,m.kackovic,
n.m.wijnberg,m.worring\}@uva.nl}
\and
Amazon AGI, United States of America
\and
College of Business and Economics,
University of Johannesburg, South Africa
}

\maketitle

\begin{abstract}
  Recent vision-language models demonstrate impressive general visual understanding, yet their art interpretation remains shallow: they describe surface content but struggle with formal analysis, grounded historical interpretation, or affective characterization. We argue this is not only a model but also a dataset limitation. Existing art datasets are single perspective resources, where no dataset provides narrative, formal, emotional, and historical perspectives simultaneously for the same artworks. We introduce MMArt, a large-scale dataset of 74,234 WikiArt paintings, each annotated with four independently annotated perspectives plus a harmonized unified caption, produced by specialized vision-language models or human annotation and validated through complementary quality evaluations. Two complementarity analyses establish that perspectives encode genuinely distinct information. A generative analysis shows that formal analysis descriptions best preserve compositional style, and historical descriptions carry strong affective signal in reconstructed images. A discriminative retrieval analysis reveals task-asymmetry: narrative descriptions drive retrieval (R@1 = $44.0\%$), while formal descriptions, strongest for reconstruction, are nearly non-discriminative at retrieval scale (R@1 = $7.8\%$). Leave-one-out analysis further confirms that historical descriptions are the least replaceable perspective across both tasks. Together, the two analyses establish that no single perspective suffices for all tasks, directly motivating MMArt's multi-perspective design. The dataset, code, and additional information are available at \url{https://shuaiwang97.github.io/MMArt/}.
  
\end{abstract}

% 1. Define the problem
% 2. General landscape: limitations of current single perspective dataset → motivation for MMArt
% 4. Our MMArt dataset and its inforamtion
% 5. how does our dataset behave and what is its analysis like

\section{Introduction}
Understanding a painting is not a simple act of perception. It is a layered process that draws simultaneously on visual attention~\cite{riccio2026genderartifactsarthistory,li2021paint4poemdatasetartisticvisualization,vannoord2025iconicitygeneratedimage}, affective response~\cite{BMVC_style_2013,panofsky1955meaning}, formal knowledge~\cite{schaerf2026artsemanticssheafinformedcontrastive,baxandall1988painting,proto_HGNN}, and historical memory~\cite{wang2026mar,ArtRAG_MM2025}. A viewer encountering Vermeer's \emph{Girl with a Pearl Earring}\footnote{\url{https://en.wikipedia.org/wiki/Girl_with_a_Pearl_Earring}} may first register the luminous skin and direct gaze: a narrative observation. A trained eye may then notice the sfumato technique and restricted palette: a formal judgment. The same viewer may feel unease or intimacy: an emotional response. And an art historian will situate the work within the Dutch Golden Age domestic portrait tradition: a contextual interpretation. These four acts of understanding are  concurrent and interacting perspectives on a single visual object.

\begin{figure}[t]
    % \vspace{-3mm}
\includegraphics[
        width=\linewidth,
        trim={6mm 0 6mm 0},
        clip
    ]{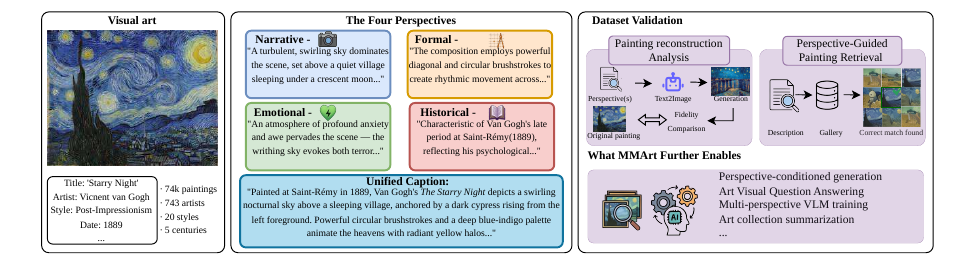}
    % \vspace{-7mm}
    \caption{Overview of the MMArt dataset. Each painting is annotated with four source-specialized perspectives: Narrative and Scene Interpretation, Formal Visual Analysis, Emotional Response, and Historical and Contextual Analysis, plus a harmonized unified caption. Perspective validity is evaluated through a text-to-image reconstruction and perspective guided retrieval experiment measuring fidelity.}
\label{fig:MMArt_overview}
\end{figure}

Recent vision-language models have demonstrated impressive capabilities in general visual understanding~\cite{NEURIPS2023_llava,chen2023sharegpt4v,bai2025qwen3vltechnicalreport,shen2026spectralprogressivethoughtflowlightweight,zeng2026sam3,wang2022replication}, yet their performance on art interpretation remains shallow: they describe surface content competently but struggle to produce formal analysis, grounded historical interpretation, or affective characterization~\cite{MM24GalleryGPT,Yuan2023ArtGPT4TA,art_icmr_2026,scala_hgnn_2024,li2024high}. Existing computational art datasets are single-perspective resources, each capturing one interpretive dimension in isolation (Tab.~\ref{tab:dataset_comparison}): SemArt~\cite{semart_2018} provides iconographic and historical commentary; ArtEmis~\cite{achlioptas2021artemis} covers affective response; ExpArt~\cite{ExpArt_acl2024} contributes expert formal analysis for a small subset; OmniArt~\cite{2017omniart} offers broad metadata coverage but no natural-language perspectives. No existing resource provides multiple interpretive dimensions simultaneously for the same artworks. As a result, models trained on these resources learn to \emph{describe} paintings, but not to \emph{interpret} them in the full sense.

We introduce \textbf{MMArt}, a large-scale multi-perspective dataset of 74,234 paintings designed to close this gap. Each painting is annotated with four independently annotated perspectives: Narrative and Scene Interpretation, Formal Visual Analysis, Emotional Response, and Historical and Contextual Analysis, plus a harmonized unified caption. A key design principle is perspective specialization: ensuring each perspective reflects genuine domain expertise rather than a single generalist model output. MMArt uses GalleryGPT~\cite{MM24GalleryGPT} for formal analysis, ArtRAG~\cite{ArtRAG_MM2025} for historically grounded context, and ArtEmis annotation-conditioned generation~\cite{achlioptas2021artemis} for emotional response. Perspectives are collected independently to preserve interpretive diversity and harmonized into a unified caption through an explanation synthesis step.

We validate MMArt's multi-perspective design through two complementarity analyses probing perspective distinctiveness from opposite directions. A \textit{generative complementarity analysis} tests whether each perspective recovers a distinct visual dimension: given only text, we measure which perspectives best reconstruct style, composition, and affective tone via text-to-image generation.  A \textit{discriminative complementarity analysis} evaluates each perspective(s) as a retrieval query against the full 74k gallery. Perspective contributions prove task-asymmetric: the narrative perspective drives retrieval (R@1\,=\,44.0\%), while formal descriptions, which achieve the highest generative fidelity, are nearly non-discriminative at retrieval scale (R@1\,=\,7.8\%). Together, they establish that no single perspective suffices for all tasks, directly motivating MMArt's multi-perspective design. MMArt is designed to support perspective-conditioned model training, affective computing on visual art, knowl\-edge-intensive art question answering, and retrieval-augmented generation tasks that current single perspective datasets cannot support simultaneously. Our contributions are:

\begin{itemize}
    \item \textbf{Multi-Perspective Visual Art Dataset.} We introduce MMArt, a large-scale dataset of 74,234 WikiArt paintings each annotated with four independently generated interpretive perspectives of narrative, formal, emotional, historical, via specialized human annotation or vision-language models, and a harmonized unified caption.

    \item \textbf{Perspective Complementarity Analyses.} We propose a two-directional evaluation framework for multi-perspective art datasets: a \textit{generative} perspective-to-painting reconstruction and a \textit{discriminative} perspective-guided cross-modal retrieval, jointly probing whether perspectives encode distinct and complementary visual information.

    \item \textbf{Public Release.} MMArt is fully open: dataset, generation and validation pipeline are publicly available to ensure reproducible research in multimedia art understanding.

\end{itemize}

\section{Related Work}
We discuss related work along two key relevant dimensions: existing visual art understanding datasets and methods for multi-perspective art explanation.

\subsection{Visual Art Understanding Datasets}

Existing datasets for computational art understanding are predominantly single-perspective resources, each typically capturing a single interpretive dimension.
SemArt~\cite{semart_2018} provides  Web Gallery of Art (WGA) paintings paired with museum-catalog commentaries that blend iconographic and historical content, enabling cross-modal retrieval but offering no formal, emotional, or narrative perspective.
ArtEmis~\cite{achlioptas2021artemis} collected 439K crowd-sourced affective utterances across WikiArt paintings — the largest coverage of emotional response to art — but contains no formal, historical, or scene-level descriptions.
ExpArt~\cite{ExpArt_acl2024} contributes expert formal and historical explanations for approximately 3,500 artworks, while Artpedia~\cite{2019_Artpedia} separates visual sentences from contextual ones for 2,930 paintings.
ArtCaps~\cite{ArtCaps} and earlier captioning datasets~\cite{MM_19_art,Iconographic_IC} target general or iconographic descriptions at smaller scale.
OmniArt~\cite{2017omniart} offers the broadest coverage (432K artworks) but provides only structured metadata rather than natural-language perspectives. For visual question answering on art, AQUA~\cite{garcia_dataset_2020} and ArtQuest~\cite{bleidt_artquest_2024} introduce benchmarks that probe factual and semantic understanding but again operate within a single question-answering register.
The common limitation of these datasets is that each captures one slice of art interpretation: studying the interplay of formal analysis and affective response typically requires combining incompatible datasets with different scales, annotation conventions, and artwork populations.
MMArt addresses this gap by providing four complementary perspectives narrative, formal, emotional, and historical  simultaneously for the same 74,234 paintings, with a unified caption integrating four explicitly defined perspectives.
Table~\ref{tab:dataset_comparison} summarizes the interpretive coverage of existing datasets.

\begin{table}[h]
\centering
\caption{Comparison of art understanding datasets across interpretive dimensions: Narr. = Narrative, Form. = Formal Analysis, Emot. = Emotion, Hist. = Historical Context. \checkmark\,= dedicated annotation; $\sim$\,= partial or metadata only; --\,= absent.}
\vspace{-2mm}
\label{tab:dataset_comparison}
\setlength{\tabcolsep}{4pt}
\begin{tabular}{lrcccccc}
\toprule
\textbf{Dataset} & \textbf{Size} & \textbf{Narr.} & \textbf{Form.} & \textbf{Emot.} & \textbf{Hist.} & \textbf{Unified} \\
\midrule
SemArt~\cite{semart_2018}               & 21K  & \checkmark & $\sim$         & --         & \checkmark & -- \\
ArtEmis~\cite{achlioptas2021artemis}    & 80K  & --         & --         & \checkmark & --     & -- \\
Artpedia~\cite{2019_Artpedia}           & 3K   & \checkmark & --         & --         & $\sim$     & -- \\
ExpArt~\cite{ExpArt_acl2024}            & 3.5K & --         & \checkmark & --         & \checkmark & -- \\
ArtCaps~\cite{ArtCaps}                  & 4K   & \checkmark & --         & --         & --        & -- \\
OmniArt~\cite{2017omniart}              & 432K & --         & --         & --         & $\sim$    & -- \\
\midrule
\textbf{MMArt (ours)}                 & \textbf{74K} & \checkmark & \checkmark & \checkmark & \checkmark & \checkmark \\
\bottomrule
\vspace{-8mm}
\end{tabular}
\end{table}

\subsection{Perspective-conditioned Art Explanation}
Prior work has addressed art explanation from individual 
interpretive perspectives. Bai et al.~\cite{Bai2021ExplainMT} come closest, generating separate content, form, and context descriptions augmented with Wikipedia knowledge retrieval. However, their framework is trained on SemArt~\cite{semart_2018}, whose museum-catalog commentaries blend multiple topics without dedicated per-perspective annotation — with more than 80\% 
of paintings missing at least one perspective.
GalleryGPT~\cite{MM24GalleryGPT} fine-tunes a vision-language model for formal compositional analysis on the PaintingForm dataset; ArtGPT-4~\cite{Yuan2023ArtGPT4TA} targets general artistic understanding via an LLaVA adapter. Both confirm that specialized models outperform generalists on individual perspectives, yet neither combines multiple perspectives into a unified dataset. Knowledge grounding is equally critical for historical interpretation, where pure visual inference leads to frequent factual confabulation~\cite{llm_Hallucination_2023}. ArtRAG and VL-KGE~\cite{ArtRAG_MM2025,efthymiou2026vl} demonstrates that art-historical knowledge graphs substantially improve factual accuracy and interpretive depth. The key departure of MMArt from all prior work is scale  and coverage: where existing resources capture one interpretive dimension independently, MMArt provides all four simultaneously for 74k paintings, enabling the cross-perspective complementarity analyses that form the core of our validation.

\begin{figure*}[ht]
    \centering\includegraphics[width=\linewidth]{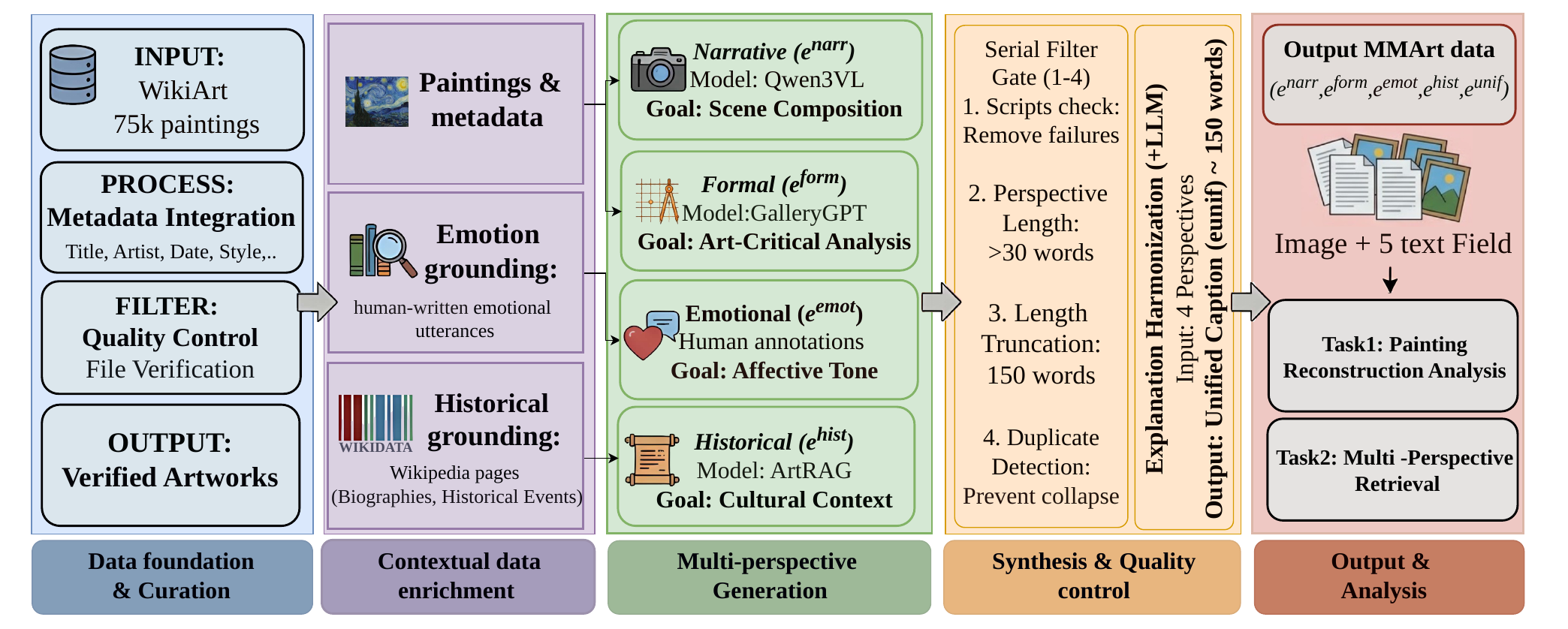}
    \vspace{-5mm}
    \caption{Overview of the MMArt dataset construction pipeline. Each painting is processed through four specialized vision-language models to produce independently annotated perspectives (narrative, formal, emotional, historical), which are then harmonized into a unified caption.}
\label{fig:MMArt_framework}
\vspace{-2mm}
\end{figure*}

\section{MMArt Dataset Construction}

The four-perspective architecture is grounded in Panofsky's framework~\cite{panofsky1955meaning} for layered art interpretation, which distinguishes factual description, formal analysis, and symbolic meaning as distinct and non-reducible acts of understanding. We operationalize this as four computational perspectives: narrative description, which supports scene-level retrieval and VQA; formal analysis, which enables style transfer and compositional modeling; emotional response, which grounds affective computing research; and historical context, which supports knowledge-intensive generation and RAG benchmarks. Each perspective targets a distinct downstream task family that no existing single-perspective dataset can support simultaneously — this is the core design rationale for MMArt.
MMArt is built on WikiArt\footnote{\url{https://www.wikiart.org}}, a publicly accessible repository that has served as the foundation for numerous computational art understanding benchmarks~\cite{Thanos_MM_21,achlioptas2021artemis,proto_HGNN}. It covers approximately 75k artworks across 20 style categories spanning the 15th through the 21st century, each associated with structured metadata including title, artist name, production date, style, and school. The remainder of this section describes the four-perspective architecture, generation pipeline, and quality control procedures.

\subsection{Four-Perspective Architecture}
Rather than generating a single descriptive caption, MMArt decomposes each artwork's annotation into four independently generated perspectives and one unified description:

\begin{equation}
P(i) = \{ e^{\text{narr}},\ e^{\text{form}},\ e^{\text{emot}},\ e^{\text{hist}} \, e^{\text{unif}} \},
\end{equation}

where $i$ denotes a painting and each component captures a distinct interpretive dimension.
\textit{Narrative and Scene Interpretation ($e^{\text{narr}}$):}
A factual account of the depicted entities, figures, scene composition, and visual elements as they appear in the image.
This perspective answers \textit{what} is shown, without invoking symbolic or historical interpretation.
\textit{Formal Visual Analysis ($e^{\text{form}}$):}
An analysis of the painting's compositional structure, including spatial organization, color palette, brushwork, use of light and shadow, and visual rhythm. This perspective corresponds to the vocabulary of formal art criticism, focusing on \textit{how} the work is constructed.
\textit{Emotional Response ($e^{\text{emot}}$):}
An affective characterization of the painting's perceived mood, atmosphere, and psychological tone by a viewer. This perspective captures the phenomenological dimension of art encounter---\textit{what it feels like} to look at the work.
\textit{Historical and Contextual Analysis ($e^{\text{hist}}$):} An art-historical interpretation situating the work within its cultural context: movement affiliation, iconographic codes, period-specific symbolic meaning, and relevant biographical or historical circumstances.
This perspective answers \textit{why} the work looks and means as it does.

In addition to the four independent perspectives, each painting receives a \textit{unified caption} $e^{\text{unif}}$ produced by harmonizing all four into a single coherent description.
This unified caption serves as a strong single-text baseline in benchmark evaluations.

\subsection{Annotation Pipeline}
A central design principle of MMArt is perspective specialization: each perspective is generated by a model chosen for its demonstrated strength in the corresponding interpretive task. 
% Generating all perspectives with a single general-purpose model risks homogenizing the output; using specialized models encourages each perspective to reflect genuine domain fit.
This is a deliberate data construction strategy, not a fragmentation of interpretation. Generating perspectives independently with specialized models ensures each dimension reflects genuine domain expertise, providing the clean per-perspective supervision that joint multi-perspective VLM training requires.

\paragraph{Narrative perspective.}
Qwen3-VL-8B-Instruct~\cite{bai2025qwen3vltechnicalreport} is selected for its broad visual pretraining that minimizes art-domain bias, keeping output anchored in observable scene content. Narrative captions are generated from the painting image and metadata alone using a prompt ($\pi_{\text{narr}}$) that explicitly prohibits symbolic, historical, or emotional reference, constraining output to visually observable entities, figures, and spatial relationships.

\paragraph{Formal analysis.}
Formal perspectives are generated using specialized GalleryGPT~\cite{MM24GalleryGPT}, a LLaVA-7B~\cite{NEURIPS2023_llava} model fine-tuned on its PaintingForm dataset specifically for formal art analysis. No general-purpose VLM matches GalleryGPT's adherence to formal criticism vocabulary on this task. Its supervised fine-tuning on expert-annotated PaintingForm data makes it the only publicly available model purpose-built for this perspective. The prompt ($\pi_{\text{form}}$)instructs the model to analyze compositional structure, palette, brushwork, and spatial organization while suppressing narrative or interpretive content.

\paragraph{Emotional response.}
Emotional perspectives are generated using Qwen3-VL-8B-Instruct~\cite{bai2025qwen3vltechnicalreport}, conditioned on the painting image together with human-written affective utterances from ArtEmis~\cite{achlioptas2021artemis,mohamed2022okay}, a large-scale emotion dataset collected over the same WikiArt corpus, providing on average $5.7$ authentic human responses per painting. Unlike ArtEmis, which provides fragmented affective utterances, MMArt synthesizes these viewer responses into coherent emotional characterizations aligned with narrative, formal, and historical perspectives for the same paintings. This ensures that affective grounding draws on real viewer reactions to the exact paintings in MMArt rather than out-of-domain sentiment signals. Qwen3-VL is chosen here because its instruction-following capability allows reliable conditioning on the ArtEmis utterances as affective anchors. The prompt ($\pi_{\text{emot}}$) instructs the model to characterize perceived mood and psychological tone grounded in the provided human utterances, without describing scene or historical context.

\paragraph{Historical and contextual analysis.}
Historical perspectives are generated using ArtRAG~\cite{ArtRAG_MM2025} augmented with structured art-historical context retrieved via its art context knowledge documents. For each artwork, the top-5 context documents covering artist biography, movement affiliation, and relevant historical events are retrieved by embedding ranking and concatenated as generation context. This retrieval-augmented approach is chosen over direct VLM generation because historical and biographical facts are not recoverable from visual appearance alone~\cite{ArtRAG_MM2025}. Instead, grounding generation in verified external knowledge substantially reduces factual error. The prompt ($\pi_{\text{hist}}$) instructs the model to situate the work within its cultural and art-historical context using the retrieved documents, explicitly suppressing scene description and subjective affect.

\paragraph{Unified caption.}
Given the four independently generated perspectives $\{e^{\text{narr}}, e^{\text{form}}, \\ e^{\text{emot}}, e^{\text{hist}}\}$, a unified caption $e^{\text{unif}}$ is produced via Qwen3-8B~\cite{bai2025qwen3vltechnicalreport} using a harmonization prompt that synthesizes the four perspectives into a single coherent, non-redundant description of approximately 150 words.
The unified caption preserves interpretive breadth while eliminating cross-perspective repetition, and is designed for downstream tasks such as retrieval and language-model grounding where a holistic single-text representation is preferred.
Formally, the generation of perspective $v$ for painting $w$ with image $I$, metadata $M$, and optional retrieved context $\mathcal{S}$ is:

\begin{equation}
e^{v} = \text{VLM}_v\!\left(I,\ M,\ \mathcal{S},\ \pi_v\right),
\end{equation}

where $\pi_v$ is a perspective-specific prompt template designed to enforce interpretive focus and minimize cross-perspective overlap. All perspectives are generated with model's default decoding temperature and max\_tokens = 256. Full prompt templates for all five generation steps are provided in the supplementary website.

\subsection{Quality Control and Data Validation}
\label{sec:quality}

\paragraph{Text quality filtering.}
We apply four automated cleaning steps to the raw generated output.
(i)~\textit{Script contamination}: perspectives containing non-Latin characters are filtered out, as these indicate generation failure in which the model switches output language rather than describing the painting.
(ii)~\textit{Length filtering}: $e$ length under 30 words are nulled and filtered out as uninformative, and explanations exceeding 150 words is truncated at the nearest sentence boundary to remove generation runoff without discarding content.
(iii)~\textit{Duplicate detection}: exact-duplicate descriptions values across paintings are nulled, as these indicate model collapse on visually similar inputs.
The full dataset retains all 75,336 paintings with nulls preserved; the experiment-ready subset requires all four perspectives to be non-null, yielding 74,234 paintings. A core claim of MMArt is that the four perspectives encode distinct information rather than stylistic paraphrases of each other. We validate this at two levels.

\paragraph{Semantic distinctiveness.}
To quantify the degree of overlap between perspectives for the same painting, we compute pairwise cosine similarity between perspective embeddings using CLIP ViT-L/14 text encodings on a random sample of 1,000 paintings.
Table~\ref{tab:pairwise_sim} reports the mean pairwise similarity between all perspective pairs.
All pairs exhibit low cosine similarity ($<$0.55), with the narrative--formal pair showing the greatest divergence and narrative--historical showing the most overlap, consistent with the shared factual grounding of both perspectives.
These results confirm that the perspectives are semantically non-redundant.

\begin{table}[h]
\centering
\vspace{-2mm}
\caption{Mean pairwise CLIP-text cosine similarity across perspectives. Lower indicate greater semantic distinctiveness.}
\vspace{-3mm}
\label{tab:pairwise_sim}
\begin{tabular}{lcccc}
\toprule
 & $e^{\text{narr}}$ & $e^{\text{form}}$ & $e^{\text{emot}}$ & $e^{\text{hist}}$ \\
\midrule
$e^{\text{narr}}$ & 1.00 & 0.41 & 0.48 & 0.52 \\
$e^{\text{form}}$ & 0.41 & 1.00 & 0.43 & 0.44 \\
$e^{\text{emot}}$ & 0.48 & 0.43 & 1.00 & 0.46 \\
$e^{\text{hist}}$ & 0.52 & 0.44 & 0.46 & 1.00 \\
\bottomrule
\vspace{-9mm}
\end{tabular}
\end{table}

\begin{table}[h]
\centering
\caption{LLM-as-judge quality ratings (1--5 scale, mean$\pm$std.)}
\label{tab:quality}
\vspace{-3mm}
\setlength{\tabcolsep}{4pt}
\begin{tabular}{lccc}
\toprule
\textbf{Perspective} & \textbf{Fidelity} & \textbf{Accuracy} & \textbf{Depth} \\
\midrule
Narrative  & 4.49$\pm$0.80 & 4.33$\pm$0.95 & 3.77$\pm$0.45 \\
Formal     & 3.70$\pm$1.17 & 3.62$\pm$1.32 & 3.03$\pm$0.74 \\
Emotional  & 4.99$\pm$0.12 & 4.41$\pm$0.71 & 3.98$\pm$0.17 \\
Historical & 4.86$\pm$0.42 & 4.28$\pm$0.76 & 3.78$\pm$0.43 \\
Unified    & 4.64$\pm$0.68 & 4.13$\pm$1.04 & 4.08$\pm$0.43 \\
\bottomrule
\vspace{-8mm}
\end{tabular}
\end{table}

\begin{figure}[t]
    \includegraphics[width=\linewidth]{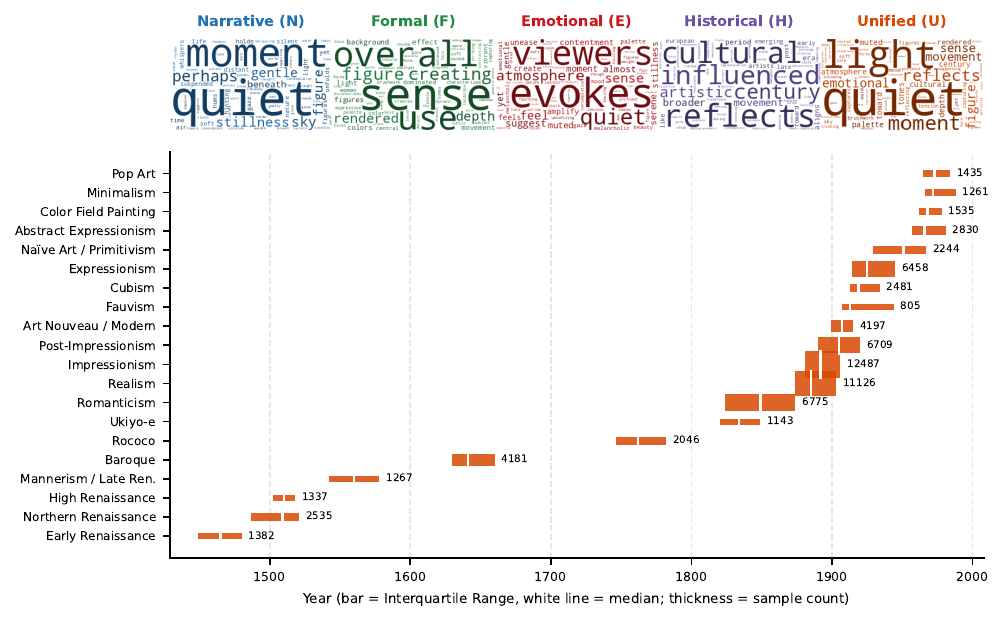}
    \vspace{-7mm}
    \caption{Top: Characteristic vocabularies for different  perspectives, using TF-IDF weighted word clouds. Bottom: Chronological distribution and volume of art styles.}
    \label{fig:dataset_overview}
    \vspace{-5mm}
\end{figure}

\paragraph{LLM-as-judge evaluation.}
We evaluate all five perspectives using Gemma-3-27B~\cite{gemmateam2025gemma3technicalreport} as an independent vision-language judge on 300 randomly stratified paintings. As a model from a different family than all generation models used in dataset construction, Gemma-3-27B mitigates self-evaluation bias.
Each perspective is rated on three dimensions (1--5 scale): \textit{perspective fidelity} (if the text focuses on the correct aspect), \textit{factual accuracy} (if the content is correct, assessed against both the painting image and metadata), and \textit{depth} (if it provides substantive detail beyond generic description). Each perspective is rated independently in a single zero-shot pass; the rating prompt instructs the judge to assess only the specified dimension without reference to other perspectives. The full judge prompt is provided in the supplementary materials.
Table~\ref{tab:quality} reports the results. Three patterns are notable. First, $e^{\text{emot}}$ achieves near-perfect fidelity (4.99$\pm$0.12) with the lowest variance of any perspective, reflecting the accurate information of ArtEmis affective labels. 
Second, $e^{\text{form}}$ receives lower scores, especially in depth (3.03$\pm$0.74), reflecting the broader valid explanation space of formal analysis, where composition, brushwork, palette, and style can be emphasized in different ways. This does not imply low utility: later reconstruction results show that formal descriptions best preserve style and composition.
Third, $e^{\text{unified}}$ achieves the highest depth score (4.08$\pm$0.43) of any perspective, confirming that the harmonization step produces captions with richer substantive detail than any single-perspective source. Figure.~\ref{fig:dataset_overview} reports the data distribution of MMArt.

\paragraph{Ethical Considerations and License.} MMArt is built entirely from WikiArt, a publicly accessible repository of artworks whose copyright has expired or which have been made available for educational use. The dataset contains no personal data, private images, or crowd-sourced human annotations beyond the published ArtEmis dataset~\cite{achlioptas2021artemis}, which was collected under institutional review. All generated perspectives describe artworks, not individuals. No licence terms of the source materials are violated. MMArt is released under the Creative Commons Attribution 4.0 International license (CC BY 4.0) for academic and research use.

\begin{figure*}[t]
    \includegraphics[width=\linewidth]{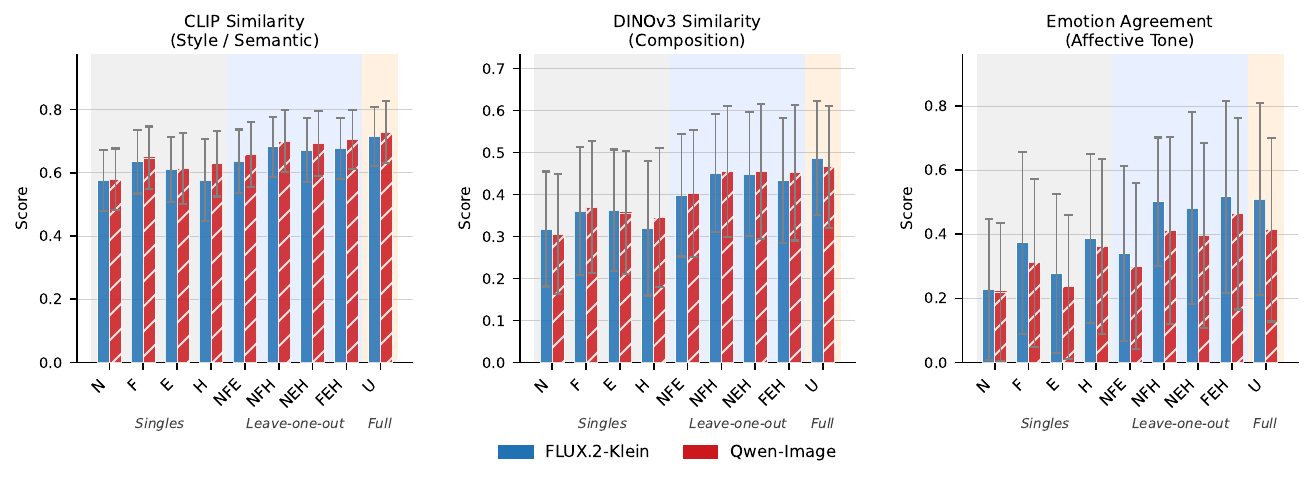}
    \vspace{-8mm}
    \caption{Reconstruction fidelity across nine perspective conditions for FLUX.2-Klein (blue) and Qwen-Image (red). Shaded regions separate singles, leave-one-out triples, and the full four-perspective condition. Error bars show $\pm 1$ standard deviation.}
    \vspace{-5mm}
    \label{fig:conditions_bar}
\end{figure*}

\section{Perspective Complementarity Analyses}
Having established that the four perspectives are 
semantically non-redundant in Section~\ref{sec:quality}, a deeper question remains: Do they contribute differently across tasks? A perspective may encode distinct information yet still be uninformative for a particular downstream application. We introduce a two-directional analysis framework to probe precisely this: whether each perspective's utility is task-asymmetric. Both analyses evaluate nine conditions: four singles ($e^\text{narr}$, $e^\text{form}$, $e^\text{emot}$, $e^\text{hist}$), four leave-one-out triples (each omitting one perspective), and the unified caption 
$e^\text{unif}$. Conditions are evaluated on the same stratified 1,000 painting samples.

\subsection{Generative and Discriminative Evaluation}

We designed two-directional analysis framework for the experiment. The generative analysis investigates whether a perspective can \emph{reconstruct} a painting, which probing what visual information each perspective encodes. 
We use reconstruction not as an artwork restoration task, but as a diagnostic probe of which visual, compositional, and affective cues each textual perspective preserves.
Given text from one or more perspectives without the original image, a text-to-image model generates a reconstruction of the original painting. Multi-perspective conditions are first synthesized  into a single coherent $\sim$80-word prompt using Qwen3-8B, ensuring all conditions enter the image generator in the same text format so that differences in fidelity are attributable to perspective content rather than formatting. Two text-to-image  models including FLUX.2-Klein-4B~\cite{flux2} and Qwen-Image-2512~\cite{wu2025qwenimage} are used to confirm that findings are model-agnostic. Reconstruction fidelity is measured against the original painting across three  complementary metrics:
\vspace{-2mm}
\begin{equation}
\mathcal{F}(w, \hat{w}) = \bigl(\delta_{\text{style}},\ 
\delta_{\text{comp}},\ \delta_{\text{emot}}\bigr),
\end{equation}

where $\delta_{\text{style}}$ is CLIP ViT-L/14~\cite{clip_2021} cosine similarity  and $\delta_{\text{comp}}$ is DINOv3-vitl16-pretrain-lvd1689m embedding cosine similarity. The two metrics are complementary by design: CLIP's image--text contrastive training makes it sensitive to global style and semantic content, while DINOv3's self-supervised image-only training yields representations more attuned to local spatial structure and compositional layout~\cite{simoni2025dinov3}. Together they capture different visual dimensions of reconstruction fidelity that a single metric would conflate~\cite{Tong_2024_CVPR}.  $\delta_{\text{emot}}$ is CLIP zero-shot emotion agreement, measuring the proportion of  paintings for which the regenerated image and the original share the same top-1 ArtEmis emotion label.

The discriminative direction asks whether a perspective can \emph{identify} a painting among all MMArt candidates, probing how unique the encoded information is. A perspective that reconstructs well may still fail to discriminate, and vice versa; only together do the two directions establish genuine complementarity. We embed all 74,234 MMArt paintings as a fixed gallery using Qwen3-VL-Embedding-2B~\cite{qwen3vlembedding} and 
Jina-CLIP-v2~\cite{jinaclipv2}, two architecturally 
distinct embedding models, to confirm that findings are model-agnostic. Given a textual perspective as a query, we evaluate description to painting retrieval on the same stratified 1,000-painting sample and nine perspective conditions. Retrieval performance is measured by Recall at $k$ (R@$k$ for $k \in \{1, 5, 10\}$), Precision at $k$ (P@$k$ for $k \in \{1, 5, 10\}$) and NDCG, Mean Reciprocal Rank (MRR), all computed over the full 74k painting gallery.

\textit{Painting Reconstruction Analysis Results.}
Figure~\ref{fig:conditions_bar} reports reconstruction fidelity across all nine conditions for image generators. The unified perspective condition $e^{\text{unif}}$ outperforms all single perspectives on both CLIP style similarity and DINOv3 composition, confirming that combining perspectives yields richer visual grounding than any individual caption. Among singles, $e^{\text{form}}$ achieves the highest CLIP fidelity, while FEH leads on emotion agreement, indicating that $e^{\text{narr}}$ contributes less to affective tone than the remaining perspectives. Leave-one-out analysis confirms that $e^{\text{hist}}$ is the hardest perspective to replace, contributing the largest marginal gain across all three metrics. Together, these results show that each perspective recovers a distinct subset of fidelity dimensions, and reconstruction improves as perspectives are combined.

% \begin{table}[h]
% \centering
% \caption{Reconstruction fidelity (mean) across perspective conditions. Best single per metric in \textbf{bold}; NFEH shown for reference. Results averaged across FLUX.2-Klein and Qwen-Image.}
% \label{tab:complementarity}
% \setlength{\tabcolsep}{5pt}
% \begin{tabular}{lccc}
% \toprule
% \textbf{Condition} & \textbf{CLIP} $\uparrow$ & \textbf{DINOv3} $\uparrow$ & \textbf{Emotion} $\uparrow$ \\
% \midrule
% $e^{\text{narr}}$ (N)  & 0.578 & 0.344 & 0.224 \\
% $e^{\text{form}}$ (F)  & \textbf{0.641} & 0.401 & 0.343 \\
% $e^{\text{emot}}$ (E)  & 0.612 & \textbf{0.416} & 0.257 \\
% $e^{\text{hist}}$ (H)  & 0.603 & 0.378 & \textbf{0.674} \\
% \midrule
% NFE (no H) & 0.647 & 0.456 & 0.320 \\
% NFH (no E) & 0.691 & 0.505 & 0.457 \\
% NEH (no F) & 0.682 & 0.498 & 0.439 \\
% FEH (no N) & 0.692 & 0.479 & 0.491 \\
% \midrule
% $e^{unified} (U)$ & \textbf{0.689} & \textbf{0.503} & \textbf{0.431} \\
% \bottomrule
% \end{tabular}
% \end{table}

\begin{figure}[t]
    \includegraphics[width=\linewidth]{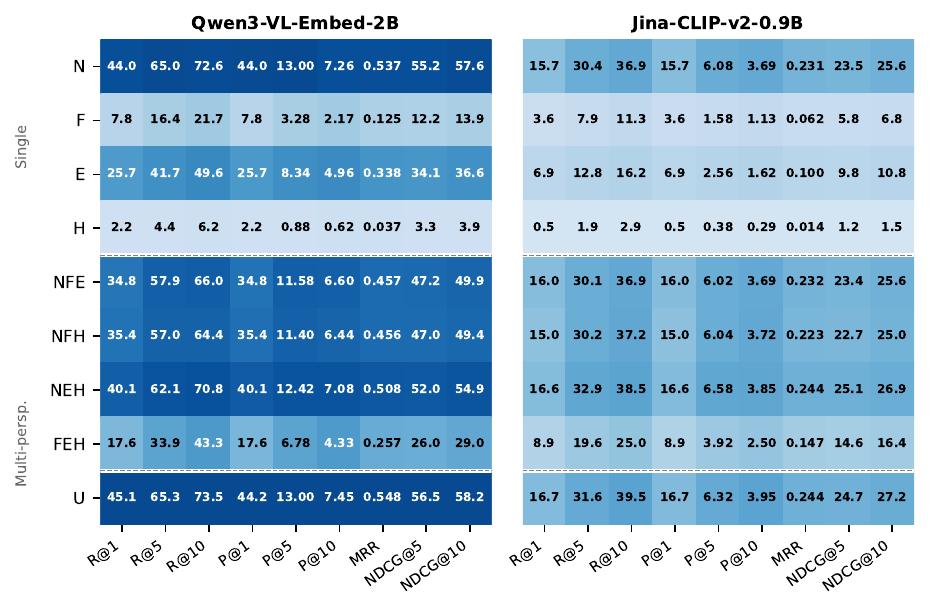}
    \vspace{-7mm}
    \caption{Description to painting retrieval performance across perspective conditions from the  full MMArt gallery. }
    \label{fig:fig_retrieval_heatmap}
    \vspace{-5mm}
\end{figure}

 \textit{Perspective-Guided Painting Retrieval Results} Figure~\ref{fig:fig_retrieval_heatmap} reports retrieval performance across all nine conditions. $e^{\text{narr}}$ is the strongest single perspective (R@1\,=\,44.0\%, MRR\,=\,0.537), while $e^{\text{hist}}$ is nearly non-functional as a retrieval query (R@1\,=\,2.2\%, MRR\,=\,0.037). 

 This does not contradict its reconstruction value: movement, period, and biographical cues are shared across many paintings and are therefore weak identifiers, but they provide complementary priors for style, composition, and affective tone.
 Interestingly, this ordering is the inverse of the regeneration results in Figure~\ref{fig:conditions_bar}, where $e^{\text{form}}$ achieves the highest CLIP fidelity and $e^{\text{narr}}$ the lowest. Leave-one-out analysis further reveals that $e^{\text{hist}}$ is the least replaceable perspective across all three fidelity metrics, while$e^{\text{narr}}$'s contribution to style and composition is partially redundant when the other three perspectives are present. The unified caption $e^{\text{unified}}$ (R@1\,=\,45.1\%, MRR\,=\,0.548) marginally outperforms $e^{\text{narr}}$ by incorporating complementary scene detail from all four perspectives.

\subsection{Qualitative examples}

\begin{figure}[ht]
    \centering
    \includegraphics[
        width=\linewidth,
        trim=0 100 0 0,
        clip
    ]{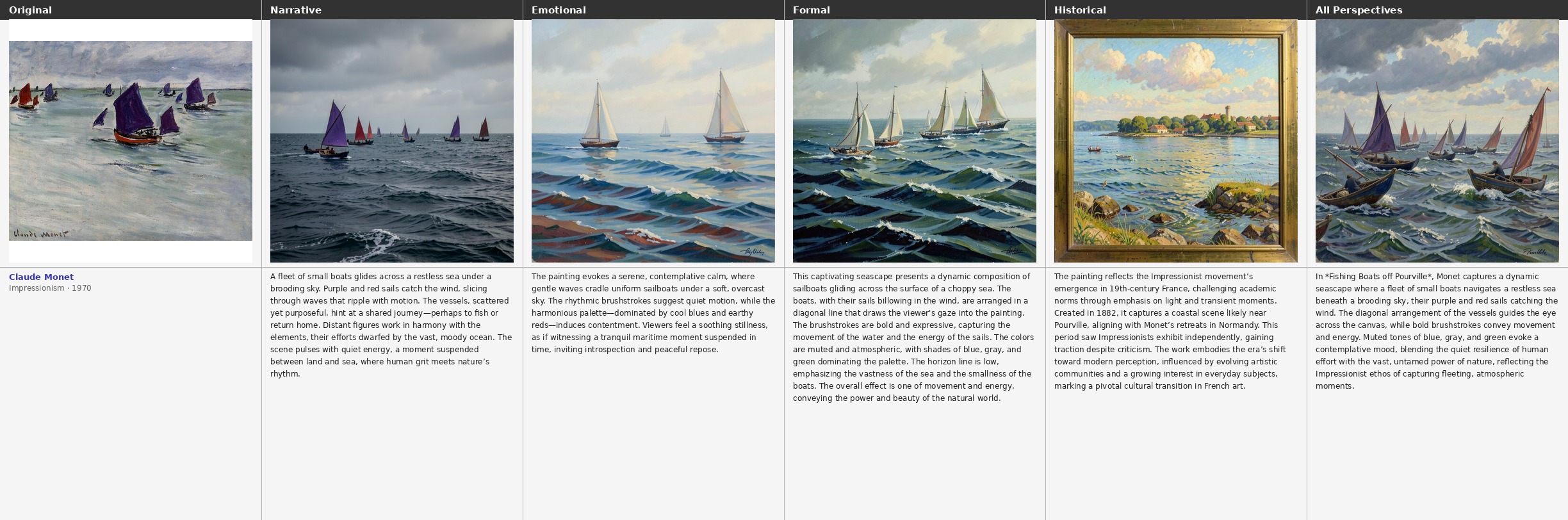}
    \caption{\textbf{Claude Monet, \textit{Fishing Boats off Pourville} (Impressionism).} The loose brushwork and coastal atmosphere are partially recovered under N and E, but the formal description best preserves the painting's hazy light and compositional structure. NFEH produces the most faithful reconstruction, capturing both style and scene.}
    \label{fig:supp_monet}
\end{figure}

\begin{figure}[ht]
    \centering
    \includegraphics[
        width=\linewidth,
        trim=0 80 0 0,
        clip]{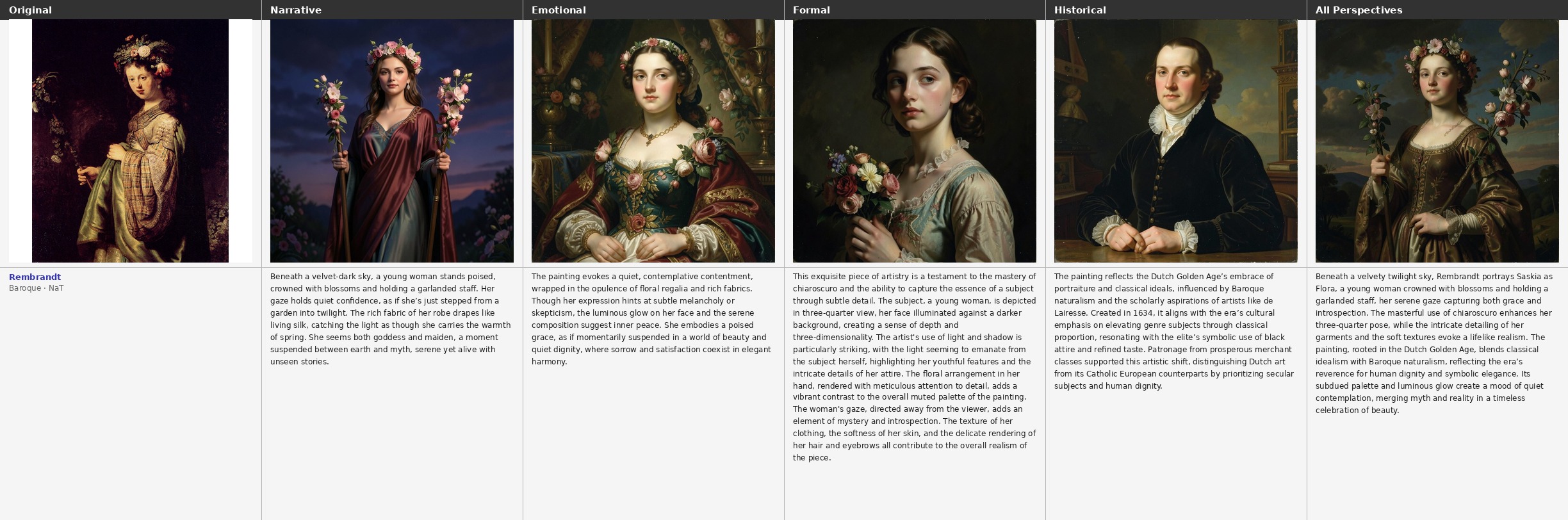}
    \caption{\textbf{Rembrandt, \textit{Saskia as Flora} (1634, Baroque).} The portrait character is strongly recovered under F, especially in chiaroscuro and three-quarter pose, and under NFEH. The historical perspective drifts toward a generic Dutch Golden Age portrait, while E captures the contemplative tone. NFEH integrates all cues into the closest match to the original.}
    \label{fig:supp_rembrandt}
\end{figure}

The Rembrandt example shows a complementary pattern for portraiture. Narrative and emotional prompts recover a flower-crowned female figure and the contemplative mood, but they only partially preserve the Baroque portrait structure. The formal prompt more strongly recovers chiaroscuro, the three-quarter pose, and the sitter's staged presentation, although it changes identity-specific details. The historical prompt introduces Dutch Golden Age portrait conventions but drifts toward a male sitter, illustrating that historical context alone can be discriminative at the level of period and genre while remaining underspecified for the depicted subject. The all-perspective reconstruction integrates the floral iconography, dark background, portrait lighting, and period cues most effectively. Together, these cases support the central claim that the four perspectives encode complementary information: narrative anchors depicted content, emotional language captures mood, formal analysis preserves composition and technique, and historical context supplies movement- and period-level priors.

\subsection{Human Evaluation}
To complement automatic metrics and LLM-as-judge scores, we conducted a blinded pairwise human evaluation against a strong general-purpose language-model baseline. We have nince annotators participated included researchers with graduate-level training in visual cultural and art understanding. 
For each trial, annotators saw the painting image and two anonymized descriptions: one from MMArt and one generated by Claude 4.5 Sonnet. The two descriptions were displayed with randomly assigned A/B labels, and annotators were not told which system produced either description.
For each criterion, annotators selected one of three options: MMArt description is better, Claude description is better, or Equal / Cannot decide.  We evaluated 25 paintings across the four criteria, yielding 100 image-perspective pairs. Nine annotators participated, and each pair received at least three independent judgments, resulting in 312 total pairwise judgments.

\begin{figure}[ht]
    \centering
    \vspace{-4mm}
    \includegraphics[width=\linewidth]{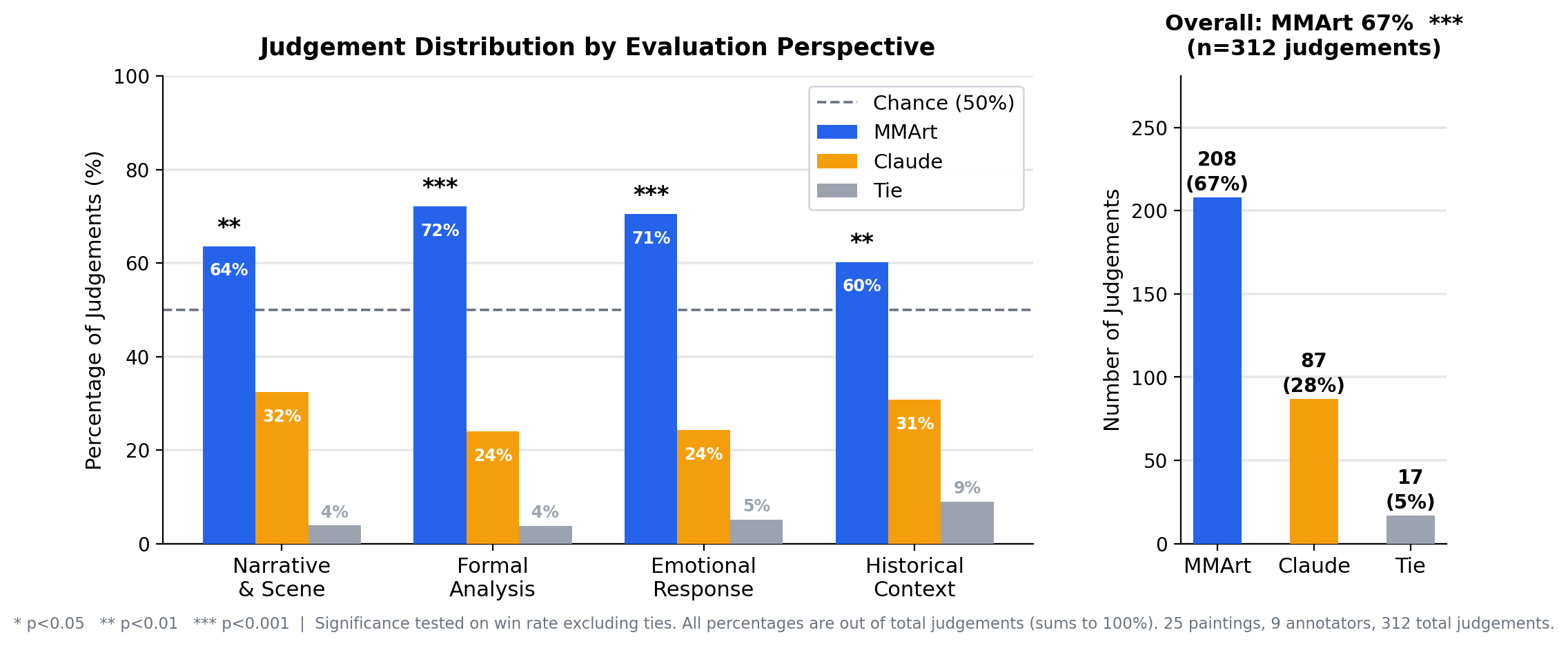}
    \caption{Human evaluation results. Left: pairwise win counts per evaluation perspective, with MMArt in blue, Claude 4.5 Sonnet in yellow, and ties in gray. Win rates and significance p-values are shown above each MMArt bar. Right: overall win rate.}
    \vspace{-6mm}
    \label{fig:supp_human_eval}
\end{figure}

Figure~\ref{fig:supp_human_eval} shows that annotators consistently preferred MMArt descriptions over the Claude 4.5 Sonnet baseline. Across all 312 judgments, MMArt was selected in 208 cases (67\%), compared with 87 selections for Claude (28\%) and 17 ties (5\%). The preference holds for every perspective: MMArt wins 64\% of narrative judgments, 72\% of formal-analysis judgments, 71\% of emotional-response judgments, and 60\% of historical-context judgments. The strongest margins occur for formal and emotional descriptions, suggesting that MMArt's perspective-specific construction produces descriptions that better match the intended interpretive criterion. Historical context is the most competitive category, but MMArt remains preferred, indicating that retrieval-grounded historical descriptions provide useful contextual specificity beyond a general-purpose baseline.

\section{Conclusion}

We presented MMArt, a large-scale dataset of 74,234 WikiArt paintings, each annotated with four independently generated interpretive perspectives and a harmonized unified caption. Two complementarity analyses establish that perspective utility is task-asymmetric. No single perspective suffices across reconstruction, retrieval, and generation, and this directly shows the effectiveness of MMArt's multi-perspective design. Leave-one-out analysis shows that historical descriptions are the least replaceable perspective across style, composition, and affective tone. The MMArt further enable perspective-conditioned VLM training, affective computing research, and knowledge-intensive art question answering and retrieval-augmented generation. Dataset, generation scripts, and pre-computed embeddings are publicly released to support reproducible research across all these applications.

\section{Acknowledgment}
This publication is part of the project The Innovation Genome, funded by the University of Amsterdam Data Science Center. This publication is also part of the project The Innovation Genome with file number EINF-14160, which is (partly) financed by the Dutch Research Council (NWO) under grant  https://doi\\.org/10.61686/EINF-14160. This work used the Dutch national e-infrastructure with the support of the SURF Cooperative.
\bibliographystyle{splncs04}
\bibliography{main}

\end{document}